\documentclass{article}
\usepackage[francais]{babel}
\usepackage[utf8x]{inputenc}
\usepackage[pdftex]{graphics}
\usepackage[T1]{fontenc}
\usepackage{amsmath,amsthm,amsfonts,url,a4wide}
\usepackage{qtree}
\def\yh#1{\ifmmode-\else\mbox{#1}\nobreak\hskip0pt\fi}
\def\monde{\ensuremath{\mathfrak{Monde}}}
\def\langue{\ensuremath{\mathfrak{LangNat}}}
\def\logique{\ensuremath{\mathfrak{LangLog}}}
\def\Trad{\ensuremath{\mathbf{Trad}}}
\def\Inter{\ensuremath{\mathbf{Inter}}}
\def\mysf#1{{\fontfamily{cmss}\selectfont\textit{#1}}} 

\newtheorem*{princo}{Principe de compositionnalité}
\newtheorem*{prin}{Principe de «composition-application»}
\newtheorem{defi}{Définition}

\title{Les mathématiques de la langue :\\ l'approche formelle de Montague\thanks{Une version plus évoluée de ce texte va paraître dans la revue \emph{Quadrature} (\texttt{http://www.quadrature.info}) en 2015.}}
\author{Yannis Haralambous}

\begin{document}
\maketitle

\begin{abstract}
Nous présentons une méthode de modélisation de la langue naturelle qui est fortement basée sur les mathématiques. Cette méthode, appelée «sémantique formelle», a été initiée par le linguiste américain Richard M. Montague dans les années 1970. Elle utilise des outils mathématiques tels que les langages et grammaires formels, la logique du 1\up{er} ordre, la théorie de types et le $\lambda$\yh-calcul. Nous nous proposons de faire découvrir au lecteur tant la sémantique formelle de Montague que les outils mathématiques dont il s'est servi.
\end{abstract}

\section{Introduction}

La modélisation mathématique de la langue que nous présentons ici, date des années 70 du siècle dernier. Les langages artificiels (comme le langage de la logique ou les divers langages de programmation) existaient déjà mais les linguistes pensaient que les langues naturelles étaient bien trop chaotiques pour que l'on puisse leur appliquer les mêmes méthodes que pour les langages artificiels. Arrive alors un certain Richard M. Montague qui prétend que c'est tout à fait possible, à condition d'utiliser des outils mathématiques évolués, et il le démontre à travers trois articles (dont le premier, de 1970, s'intitule, de manière assez provocante, \emph{L'anglais en tant que langage formel} \cite{mon70}). Malheureusement Montague est mort assez jeune\footnote{Et de manière tragique, au point où sa mort a inspiré un roman policier, la \emph{Sémantique du meurtre} d'Aifric Campbell \cite{meurtre}.} mais son travail a néanmoins révolutionné la linguistique et l'a rapproché des mathématiques et de l'informatique.

Dans cet article nous allons décrire les outils mathématiques dont s'est servi Montague pour modéliser la langue naturelle : les langages formels, les grammaires formelles, le $\lambda$\yh-calcul typé, la logique du 1\up{er} ordre, la théorie des ensembles. Ce texte s'inspire fortement (tout en le simplifiant) de l'ouvrage~\cite{cann}.

\section{L'approche de Montague}

La langue sert avant tout de parler du monde. Nos points de départ seront donc le «monde» et la «langue». Le  «monde», que nous noterons \monde{} peut être le monde réel, un monde imaginaire, ou simplement un ensemble d'objets ou d'idées abstraites (d'ailleurs, à la section~\ref{mp} nous parlerons de l'ensemble des «mondes possibles»). La «langue (naturelle)», que nous noterons \langue{}, sera un ensemble de phrases françaises, orthographiquement et syntaxiquement correctes, et qui se réfèrent aux objets de l'ensemble \monde{}.

Pour étudier la langue, Montague propose d'intercaler entre \langue{} et \monde{}, un ensemble de formules écrites dans un langage artificiel intermédiaire, basé sur la logique de 1\up{er} ordre et le $\lambda$\yh-calcul typé. Nous noterons par \logique{} l'ensemble des traductions des phrases de \langue{} dans ce langage.

On a donc la situation suivante :
\[
\langue\xrightarrow{\Trad}\logique\xrightarrow{\Inter}\monde,
\]
où \Trad{} est la \emph{traduction} des phrases françaises en langage intermédiaire, et \Inter{} le lien entre \logique{} et le \monde{} que l'on appelle \emph{interprétation}\footnote{Attention, ici le mot «interprétation» a un sens technique strict  (cf.~\S\,\ref{inter}), qu'il ne faut pas confondre avec celui du langage courant.}.

Dans la suite nous utilisons des \mysf{caractères bâton inclinés} pour représenter les mots de \langue, et des \textbf{caractères gras} pour représenter les entités de \monde. Ainsi, $\text{\mysf{Gérard}}\in\langue$ sera un nom (propre) masculin de six lettres, qui se prononce {\font\x=tipa12\x/\char'132 e.\char'113 a\char'113 /}. Par contre, $\text{\textbf{gérard}}\in\monde$ sera un individu, en chair et en os, identifié par ce prénom. On va considérer que dans \monde{} il n'y a qu'un seul \textbf{gérard}.

Dans cet article nous allons décrire, dans l'ordre, \langue{} (\S\,\ref{langue}), \logique{}  (\S\,\ref{logic}), \Trad{} (\S\,\ref{trad}) et \monde{}/\Inter{}  (\S\,\ref{monde}). 


\section{La structure de la langue}\label{langue}

Parmi les nombreuses manières d'étudier la langue naturelle, celles qui vont nous intéresser dans cet article sont la \emph{syntaxe} et la \emph{sémantique}.

La syntaxe étudie \emph{grosso modo} la composition des phrases, c'est-à-dire la manière dont les mots se combinent pour former des phrases, selon leurs catégories grammaticales (nom, adjectif, adverbe, verbe, etc.). 

La sémantique est l'étude de la signification.

Il s'avère que la syntaxe est un outil indispensable pour l'étude de la sémantique grâce au principe suivant :

\begin{princo} La sémantique d'une phrase s'obtient à partir des sémantiques de ses parties et de la manière dont elles ont été composées (= la syntaxe de la phrase).
\end{princo}

En effet, pour comprendre le sens de la phrase \mysf{Gérard aime Alice} il faut savoir ce que sont (ou peuvent être) Gérard, Alice et l'action d'aimer, et identifier par la syntaxe de la phrase le fait que c'est Gérard qui est en position de sujet et c'est donc lui qui aime Alice, et non pas l'inverse.

Pour commencer, voyons comment analyser syntaxiquement les phrases de \langue. Les outils mathématiques qui vont nous servir sont les \emph{langages formels} et les \emph{grammaires formelles}.

\subsection{Langages formels}

Définissons d'abord la notion de \emph{monoïde libre}. Soit $\Sigma$ un ensemble quelconque que nous allons appeler \emph{alphabet}. 

\begin{defi} On appelle \textbf{monoïde libre} $(\Sigma^*,\cdot,\varepsilon)$ sur $\Sigma$, l'ensemble de tous les produits $x_1\cdots x_n$ $(n\geq1)$ d'éléments de $\Sigma$, auquel on ajoute l'élément $\varepsilon$, appelé \emph{élément neutre}. La loi $\cdot$ doit être associative.
\end{defi}

On appelle \emph{concaténation} la loi du monoïde, et on s'autorise de ne pas la noter.
Les éléments du monoïde sont appelés «mots», et $\varepsilon$ est le «mot vide».

\begin{defi}
Un \textbf{langage formel} $L$ sur un alphabet $\Sigma$ est un sous-ensemble quelconque du monoïde libre $\Sigma^*$.
\end{defi}

Ce qui fait la force des langages formels, c'est que --- malgré son appellation~---, un «alphabet» n'est pas forcément un ensemble de lettres. Les éléments d'un «alphabet» peuvent être des mots-clé d'un langage de programmation (dans ce cas, un «mot» sera, par exemple, un programme dans ce langage), des symboles mathématiques (un «mot» sera alors une formule), des nucléotides (un «mot» sera alors un sous-brin d'ADN), etc. Dans notre cas, l'«alphabet» sera formé de mots de la langue française, et les éléments de \langue{} seront des \emph{phrases} de la langue française.

Se pose alors la question : si un langage formel est un sous-ensemble quelconque du monoïde libre, alors comment le décrire ? 

Toute la difficulté est là : comment choisir les mots qui forment un langage formel --- ou, de manière équivalente, comment décider si un mot appartient ou non à un langage formel donné (éventuellement infini) ?

Une méthode pour décrire des langages formels est celle des \emph{grammaires formelles}.

\subsection{Grammaires formelles}

Une grammaire formelle est un ensemble de \emph{règles de production}. En partant d'un symbole appelé «axiome de départ», on applique ces règles à un ensemble de symboles auxiliaires appelés «non-terminaux» jusqu'à aboutir aux mots du langage que l'on souhaite définir (et qui donc s'appellent «terminaux», puisqu'on ne peut aller plus loin).

\begin{defi}
Soit $\Sigma$ un alphabet. Une \textbf{grammaire formelle} $G$ est un quadruplet $(S,V,\Sigma,P)$ où $S$ est un élément appelé \textbf{axiome de départ}, $V$ l'ensemble des \textbf{symboles non-terminaux}, $\Sigma$ l'ensemble des \textbf{symboles terminaux}, et $P$ l'ensemble des \textbf{productions}. 

Une \textbf{production} $p$ est un couple $(X,u)$, où $X\in V\cup \{S\}$, et $u\in(\Sigma\cup V)^*$ (c'est-à-dire le monoïde libre sur les éléments de $\Sigma$ et de $V$). 

Une suite d'applications successives de productions est appelée une \textbf{dérivation}.
\end{defi}

On note une telle production $p:X\to u$. 

Une production envoie donc un élément non-terminal (d'où son nom de «non-terminal») ou l'axiome de départ, à un mot comportant des terminaux et des non-terminaux.

On peut appliquer des productions aux mots : si $p:X\to u$ et si le mot $w$ s'écrit $w_1Xw_2$ alors $p$ produira le mot $w_1uw_2$. Mais attention : si $w$ s'écrit $w_1Xw_2Xw_3$ alors une application de $p$ peut produire $w_1uw_2Xw_3$ ou $w_1Xw_2uw_3$ (mais pas les deux en même temps, pour que cela arrive il faudrait l'appliquer deux fois). Et si $w$ ne contient pas $X$, alors $p$ laisse $w$ inchangé.

On se pose alors la question suivante : est-ce qu'un mot~$w$ peut être l'image d'une dérivation partant de $S$ ? On dira alors que $w$ est \emph{$G$\yh-dérivable}.

\begin{defi}
Soit $G=(S,V,\Sigma,P)$ une grammaire formelle. Le \textbf{langage formel engendré par} $G$, noté $L_G$, est l'ensemble des éléments de $\Sigma^*$ qui sont $G$\yh-dérivables.
\end{defi}

Notons bien que les éléments de $L_G$ sont tous des mots sur des \emph{terminaux}. En effet, l'axiome de départ et les non-terminaux sont indispensables pour obtenir le langage formel engendré mais n'y apparaissent pas eux-mêmes. Dans notre cas, les non-terminaux vont être des catégories et fonctions grammaticales. Prenons un exemple. On peut décrire syntaxiquement la phrase \mysf{Gérard dort} par la grammaire

\smallskip

\noindent $p_1:\text{S}\to\text{GN GV}$\\
$p_2:\text{GN}\to\text{N}$\\
$p_3:\text{GV}\to\text{V}$\\
$p_4:\text{N}\to\text{\mysf{Gérard}}$\\
$p_5:\text{V}\to\text{\mysf{dort}}$

\smallskip

\noindent où S est l'axiome de départ ; les symboles GN (groupe nominal), GV (groupe verbal), N (nom) et V (verbe) sont des non-terminaux ; et, enfin, \mysf{Gérard} et \mysf{dort} sont des terminaux.

\subsection{Arbre syntaxique d'une phrase}

Comme on vient de le voir, pour décrire la structure syntaxique d'une phrase, il suffit de donner sa dérivation partant de $S$, c'est-à-dire l'ensemble de productions à travers lesquelles on obtient cette phrase.

Cette dérivation peut être admirablement bien illustrée par un \emph{arbre syntaxique} : il s'agit d'un arbre (c'est-à-dire d'un graphe connexe sans cycle) orienté, dont la racine est l'axiome initial, les sommets intermédiaires sont les non-terminaux qui interviennent dans la production de la phrase, et les feuilles sont les mots de la phrase. Tout ensemble d'arêtes (orientées) sortant d'un sommet représente alors une production, et les enfants d'un sommet sont exactement les symboles de la partie droite de sa production.

\begin{figure}[ht]
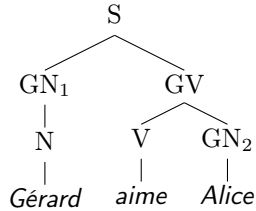

\Tree [.S [.GN$_1$ [.N \mysf{Gérard} ] ] [.GV [.V \mysf{aime} ] [.GN$_2$ \mysf{Alice} ]  ] ]
\caption{Arbre syntaxique de la phrase \mysf{Gérard aime Alice}.\label{arbre}}
\end{figure}

L'arbre syntaxique de la fig.~\ref{arbre} représente la dérivation partant de $S$ qui produit la phrase \mysf{Gérard aime Alice}.

\section{Le langage intermédiaire}
\label{logic}

La prochaine étape consiste à décrire \logique, c'est-à-dire les formules du \emph{langage intermédiaire}. Ce langage est basé sur deux outils mathématiques : la logique du 1\up{er} ordre et le $\lambda$\yh-calcul typé.

\subsection{Notions de logique du 1\up{er} ordre}

Nous ne développerons ici de la logique du 1\up{er} ordre que ce qui est nécessaire pour notre exposé.

Pour définir les \emph{formules logiques} nous allons nous servir des symboles suivants :
\begin{enumerate}
\item symboles de constante $a,b,c,\ldots$ ;
\item symboles de variable $x,y,z,\ldots$ ;
\item symboles de prédicat $P,Q,R,\ldots$ qui prennent un certain nombre d'arguments (on appelle \emph{arité} le nombre d'arguments d'un prédicat). Les arguments peuvent être des variables ou des constantes. On note, par exemple, $P(a,b,x)$ le prédicat ternaire dont les arguments sont $a,b,x$ ;
\item symboles d'opérateur qui s'appliquent aux prédicats : l'opérateur unaire de négation $\neg$, et les opérateurs binaires de conjonction $\wedge$, de disjonction $\vee$, d'implication $\to$, de double implication~$\leftrightarrow$ ;
\item symboles de quantificateur existentiel $\exists$ et universel $\forall$, qui s'appliquent aux variables ;
\item le symbole d'égalité $=$ ;
\item des parenthèses.
\end{enumerate}

\begin{defi}\label{fl}On définit une \textbf{formule logique} de manière récursive :
\begin{enumerate}
\item[\rm1.] Un prédicat d'arité $n$ appliqué à $n$ constantes et/ou variables est une formule.
\item[\rm2.] Si $a$ et $b$ sont des constantes ou des variables, $a=b$ est une formule. 
\item[\rm3.] Si $f$ est une formule, $(f)$ et $\neg f$ sont des formules.
\item[\rm4.] Si $f$ est une formule et $x$ une variable, $\exists x\,f$ et $\forall x\,f$ sont des formules.
\item[\rm5.] Si $f$ et $g$ sont des formules, $f\wedge g$, $f\vee g$, $f\to g$ et $f\leftrightarrow g$ sont des formules.
\end{enumerate}
\end{defi}

Le lecteur aura sans doute remarqué que nous venons d'énumérer une multitude de symboles sans donner leur signification.

Il y a une raison à cela : en fait, une formule logique n'est \emph{a priori} rien d'autre qu'un assemblage de symboles abstraits (selon les règles syntaxiques données par la définition~\ref{fl}) et c'est grâce à ce degré élevé d'abstraction qu'elle peut être appliquée à une infinité de situations différentes. 

Pour appliquer une formule à une situation et lui donner ainsi un «sens», on fait une \emph{interprétation}, c'est-à-dire une correspondance entre \logique{} et \monde{}. Mais qu'est-ce qu'une interprétation au juste ?

\begin{defi}
Soit $Z_2$ l'ensemble $\{\mathbf{vrai},\mathbf{faux}\}$.
Une \textbf{interprétation}\label{inter} d'une formule logique $\phi$ est une application $\Inter\colon\logique\to\monde$ telle que
pour chaque constante $c$ de $\phi$, $\Inter(c)\in\monde$, et pour chaque prédicat $n$\yh-aire $P$ de $\phi$, $\Inter(P)$ est une fonction $\monde^n\to Z_2$. D'autre part, pour toute variable~$x$ de $\phi$, $\Inter(x)$ devient une variable à valeurs dans \monde{}.
\end{defi}

L'interprétation de l'opérateur unaire $\neg$ est une application $Z_2\to Z_2$ et celles des opérateurs binaires $\wedge$, $\vee$, $\to$ et $\leftrightarrow$ sont des applications ${Z_2\times Z_2\to Z_2}$, dont les valeurs sont données par la table suivante (poétiquement appelée \emph{table de vérité}) :

%
\begin{center}{\begin{tabular}{@{}cc|c|cccc@{}}
$\phi$&$\psi$&$\neg\phi$&$\phi\wedge\psi$&$\phi\vee\psi$&$\phi\to\psi$&$\phi\leftrightarrow\psi$\\\hline
\textbf{vrai}&\textbf{vrai}&\textbf{faux}&\textbf{vrai}&\textbf{vrai}&\textbf{vrai}&\textbf{vrai}\\
\textbf{vrai}&\textbf{faux}&\textbf{faux}&\textbf{faux}&\textbf{vrai}&\textbf{faux}&\textbf{faux}\\
\textbf{faux}&\textbf{vrai}&\textbf{vrai}&\textbf{faux}&\textbf{vrai}&\textbf{vrai}&\textbf{faux}\\
\textbf{faux}&\textbf{faux}&\textbf{vrai}&\textbf{faux}&\textbf{faux}&\textbf{vrai}&\textbf{vrai}\end{tabular}}
\end{center}
%

Dans une interprétation donnée, une formule qui ne contient aucune variable est nécessairement vraie ou fausse. Il en est de même lorsqu'elle contient uniquement des variables quantifiées, c'est-à-dire telles que pour chacune il y ait un quantificateur qui lui soit appliqué. Une variable non quantifiée est appelée \emph{libre}. L'interprétation d'une formule contenant $n$ variables libres est une fonction $\monde^n\to Z_2$ : sa valeur de vérité dépend des valeurs que prennent ses variables libres.

Après cette rapide introduction à (une partie de) la logique du 1\up{er} ordre, voyons comment traduire les phrases de \langue{} en des formules logiques.

\section{Traduction $\langue \to \logique$}\label{trad}

Nous allons noter \Trad{} l'application qui traduit les phrases de \langue{} en des formules de \logique.

Une phrase du type \mysf{Gérard dort} décrit une situation où il y a un agent identifié par le nom \mysf{Gérard} qui effectue l'action de dormir. Dans la formule logique il est naturel de prendre une constante $g$ pour représenter \mysf{Gérard}.

À son tour, $g$ doit être interprétée par l'entité du monde réel qui correspond à \mysf{Gérard}, notons cette entité \textbf{gérard}. On a donc $\Trad(\text{\mysf{Gérard}})=g$ et $\Inter(g)=\text{\textbf{gérard}}$.

Le choix naturel pour traduire le verbe \mysf{dort} est un prédicat unaire $\mathrm{dort}$. L'interprétation de $\mathrm{dort}$ va être une fonction $\monde\to Z_2$, et, en particulier, on aura $\Inter(\mathrm{dort}(g))=\Inter(\mathrm{dort})(\Inter(g))=\Inter(\mathrm{dort})(\text{\textbf{gérard}})=\text{\textbf{vrai}}$.

Mais comment déduire la formule $\mathrm{dort}(g)$ à partir de la grammaire formelle de la syntaxe de la phrase \mysf{Gérard dort} ? Cette grammaire nécessite les règles suivantes :

\smallskip

\noindent $p_1:\text{S}\to\text{GN GV}$\\
$p_2:\text{GN}\to\text{N}$\\
$p_3:\text{GV}\to\text{V}$\\
$p_4:\text{N}\to\text{\mysf{Gérard}}$\\
$p_5:\text{V}\to\text{\mysf{dort}}$

\smallskip

Tous les symboles de cette grammaire sont traduits dans \logique : en effet, GV et V sont traduits par le prédicat $\mathrm{dort}$, GN et N par la constante $g$ et, par le principe de compositionnalité, S devient alors $\mathrm{dort}(g)$. On définit :

\smallskip

\noindent $\Trad(S) = \mathrm{dort}(g)$\\
$\Trad(GN) = \Trad(N) = g$\\
$\Trad(GV) = \Trad(V) = \mathrm{dort}$.

\smallskip

Notre traduction est complète puisque tous les sommets de l'arbre syntaxique de la phrase de \langue{} ont été traduits dans \logique.

\medskip

Prenons maintenant une phrase légèrement plus complexe : \mysf{Gérard aime Alice} (cf. fig.~\ref{arbre}). On peut s'attendre à avoir $\Trad(S)=\mathrm{aime}(g,a)$ où $g$ et $a$ sont des constantes logiques dont les interprétations sont le vilain \textbf{gérard} et la belle \textbf{alice}. De même, dans ce cas, $\Trad(\mysf{aime})=\mathrm{aime}$.

Mais attention ! Avons-nous traduit tous les sommets de l'arbre de la fig.~\ref{arbre} en \logique ? Hélas, non ! Car, quelle est alors la traduction de GV ? On ne peut écrire $\mathrm{aime}(\_,a)$, cela n'est pas une formule logique valide...

Montague aurait pu s'arrêter là, en disant : «je sais traduire mes terminaux (\mysf{Gérard}, \mysf{aime}, \mysf{Alice}) et mon axiome de départ ($S$) en langage logique, peu me chaut le reste».

Que nenni ! Son génie a consisté --- entre autres --- à dire que si l'on veut être honnête avec soi-même, si l'on veut aller au fond des choses, alors \emph{le principe de compositionnalité doit s'appliquer partout}, aussi bien dans \langue, que dans \logique, et la traduction d'une composition doit être la composition des traductions.  

Autrement dit : si S produit GN et GV, alors la traduction de S doit s'obtenir à partir des traductions de GN et de GV.
Mais avec les outils mathématiques décrits jusqu'à maintenant, cela n'est pas possible. Il faut donc se servir d'outils plus performants. Montague en a choisi deux : la théorie des types et le $\lambda$\yh-calcul.

Voyons d'abord ce que sont les types et à quoi ils servent.

\subsection{Théorie des types}

Avant de traduire en \logique{} tous les sommets d'un arbre syntaxique, il faut déjà se demander de quelle manière ils se combinent entre eux.

Un exemple : dans la phrase \mysf{Gérard aime Alice}, l'interprétation du prédicat binaire \mysf{aime} peut être considérée comme une application $\monde^2\to Z_2$ (elle envoie la paire d'entités $(\text{\textbf{gérard}},\text{\textbf{alice}})$
vers la valeur \textbf{vrai} si \textbf{gérard} aime \textbf{alice} et vers \textbf{faux} sinon). 

Autre exemple : dans la phrase \mysf{Gérard aime Alice et Paul déteste Virginie}, la particule de coordination \mysf{et} va combiner deux phrases pour en produire une nouvelle, son interprétation sera donc une application $Z_2\times Z_2\to Z_2$.

La situation se complique encore plus dans le cas des adverbes : dans \mysf{Gérard aime beaucoup Alice}, l'adverbe \mysf{beaucoup} agit sur le verbe, donc on peut considérer qu'il transforme une application $\monde^2\to Z_2$ en une autre application $\monde^2\to Z_2$. 

Et que dire alors des modificateurs d'adverbe comme \mysf{vraiment beaucoup} : en effet, \mysf{vraiment} agit sur \mysf{beaucoup} et est donc un transformateur de transformateur d'application $\monde^2\to Z_2$... 

Comment gérer cette complexité qui semble croître inexorablement ?

Voici la modélisation mathématique qui nous délivre du cauchemar décrit ci-dessus : Montague considère qu'il n'existe que deux \emph{types sémantiques primitifs} : celui de «formule» (dont l'interprétation dans \monde{} sera \textbf{vrai} ou \textbf{faux}) et celui de «constante individuelle» (dont l'interprétation sera un élément de \monde). Il note $t$ les formules et $e$ les constantes. Ensuite il prend le monoïde libre $\{e,t\}^*$ dont il note la loi comme un produit scalaire $\langle {,}\rangle$. Attention : cette loi \emph{n'est pas} associative, donc pas question de faire des «simplifications» : $\langle e,\langle e,t\rangle\rangle$ n'est pas la même chose que $\langle \langle e,e\rangle, t\rangle$ !

Ensuite, il appelle les éléments de ce monoïde, des \emph{types sémantiques complexes} et fait correspondre chaque sommet de l'arbre syntaxique à un type sémantique complexe, de la manière suivante : l'élément de gauche de $\langle{,}\rangle$ est la «donnée d'entrée» du type, son élément de droite est sa «donnée de sortie». 

Exemple : un prédicat unaire, comme \mysf{dort}, s'applique à une constante, son interprétation fournit une valeur de $Z_2$. On dira donc qu'il est de type $\langle e,t\rangle$ (=~«il prend un $e$ et il nous rend un~$t$»).

Mais attention : on ne prend qu'un seul élément d'entrée à la fois ! Un prédicat \emph{binaire} ne sera donc pas de type $\langle\{e,e\},t\rangle$ (cette syntaxe n'est pas valide) mais sera décrit de \emph{manière récursive} comme $\langle e,\langle e,t\rangle\rangle$. 

Si on y réfléchit un peu, c'est parfaitement logique : un prédicat binaire auquel on fournit une valeur devient prédicat unaire, donc pour une entrée $e$, la sortie est $\langle e,t\rangle$. De même, le type d'un prédicat ternaire sera $\langle e,\langle e,\langle e,t\rangle\rangle\rangle$ et ainsi de suite...

De la même manière, la particule de conjonction \mysf{et} sera de type $\langle t,\langle t,t\rangle\rangle$, l'adverbe \mysf{beaucoup} de type $\langle \langle e,t\rangle,\langle e,t\rangle\rangle$, et le modificateur d'adverbe \mysf{vraiment}, de type $\langle\langle \langle e,t\rangle,\langle e,t\rangle\rangle,\langle \langle e,t\rangle,\langle e,t\rangle\rangle\rangle$. Arrivé à ce stade, le lecteur/la lectrice doit normalement se sentir ébloui(e) devant l'époustouflante beauté de ce modèle : en effet, \emph{quelque soit} le type sémantique d'un mot, aussi complexe soit-il, on peut le décrire simplement par un élément du monoïde libre $\{e,t\}^*$... c'est simple et efficace.

\begin{figure}[h]
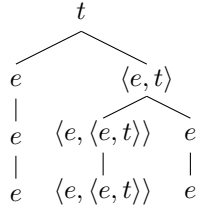

\Tree [.{$t$} [.$e$ [.$e$ $e$ ] ] [.$\langle e,t\rangle$ [.{$\langle e,\langle e,t\rangle\rangle$} {$\langle e,\langle e,t\rangle\rangle$} ] [.$e$ $e$ ]  ] ]
\caption{Types sémantiques de l'arbre de la fig.~\ref{arbre}.\label{ts}}
\end{figure}

Nous réécrivons sur la fig.~\ref{ts} l'arbre syntaxique de la fig.~\ref{arbre} en remplaçant les symboles de la grammaire formelles par leurs types sémantiques. Le lecteur peut constater que (a)~à chaque fois qu'un sommet n'a qu'un seul enfant, le type sémantique ne change pas, et (b)~à chaque fois qu'un sommet a plusieurs enfants, leurs types sémantiques se \emph{composent} : ainsi, $\langle e,t\rangle$ appliqué à $e$ donne $t$, et $\langle e,\langle e,t\rangle \rangle $ appliqué à $e$ donne $\langle e,t\rangle $.  On note $\times$ cette composition : $\langle e,t\rangle \times e=t$. Pour qu'une composition $T_1\times T_2$ puisse avoir lieu, il faut que $T_1$ soit un type complexe et que sa première composante soit égale à $T_2$.

La cohérence sémantique d'une phrase provient du fait que les types sémantiques des sommets de son arbre syntaxique se composent correctement, pour arriver au type de $S$ qui est, invariablement, $t$ (comme on peut le constater sur le graphe de la fig.~\ref{ts}).

Tout cela est bien pensé, mais on constate que l'on ne sait toujours pas comment écrire les formules logiques correspondant aux sommets intermédiaires de l'arbre syntaxique. C'est là que le $\lambda$\yh-calcul vient à la rescousse.

\subsection{Le $\lambda$\yh-calcul}

Sous ce nom exotique et mystérieux se cache tout simplement la notion de \emph{fonction} : appliquer un «$\lambda$\yh-opérateur» à une expression mathématique ou logique revient tout simplement à la transformer en fonction. 

Ainsi, $\lambda x.f(x)$ est la même chose que $x\mapsto f(x)$, c'est-à-dire la fonction~$f$. L'intérêt de la notation est qu'elle nous permet de définir toutes sortes de fonctions. Par exemple, $\lambda x.\lambda y.(x+y)$ est la fonction (de deux variables) qui à $(x,y)$ associe $x+y$, alors que $\lambda x.(x+y)$ est la fonction (d'une variable) qui à $x$ associe la somme $x+y$ (sans donner plus d'information sur $y$). De même, $\lambda P.P(x)$ est la fonction qui associe à un prédicat unaire sa valeur en $x$ et $\lambda P.\lambda x.P(x)$ est la fonction qui à $P$ et à $x$ associe $P(x)$. Les amateurs de $\lambda$\yh-calcul s'amusent même à noter $\lambda x.x$ la fonction identité et $\lambda x.c$ la fonction constante de valeur $c$.

Lorsqu'on a une fonction $f$, on peut l'appliquer à une valeur $x$, et on note le résultat $f(x)$. De même, on peut appliquer une $\lambda$\yh-expression à une valeur. Ainsi $(\lambda x.\sin(x))(\frac{\pi}{2})$ est tout simplement $\sin(\frac{\pi}{2})$. On appelle cela, tout naturellement, une \emph{application}.

L'utilisation que Montague fait de la notation $\lambda$ est très judicieuse. Elle obéit au principe suivant : 

\begin{prin}Toute \emph{composition} de types sémantiques correspond à une \emph{application} de $\lambda$\yh-expressions.\end{prin}

Ce principe va nous guider pour retrouver les $\lambda$\yh-expressions correspondant aux sommets des arbres syntaxiques. 

Prenons comme exemple la phrase \mysf{Gérard aime Alice} et les arbres des fig.~\ref{arbre} et~\ref{ts}. On sait déjà que $\Trad(\mathrm{GN}_1)=g$, $\Trad(\mathrm{GN}_2)=a$ et $\Trad(\mathrm{S})=\mathrm{aime}(g,a)$. Appliquons le principe de composition-application pour trouver $\Trad(\mathrm{GV})$ et $\Trad(\mathrm{V})$.

Mais avant de le faire, un petit changement s'impose, afin de nous mettre en conformité avec la théorie des types : on n'écrira plus $\mathrm{aime}(g,a)$ pour \mysf{Gérard aime Alice}, comme on l'a fait jusqu'à maintenant, mais $\mathrm{aime}(a)(g)$. Cela ne change en rien l'amour indéfectible de Gérard pour Alice, c'est juste que maintenant «$\mathrm{aime}$» n'est plus un «banal prédicat binaire», mais est fièrement devenu un type complexe $\langle e,\langle e,t\rangle\rangle$ !

Raisonnons maintenant à reculons : on vient de décréter que $\Trad(\text S)=\mathrm{aime}(a)(g)$ ; d'autre part, on sait que $\Trad(\text{GN}_1)=g$ ; quel sera $\Trad(\text{GV})$ ?

Laissons-nous guider par les types : dans la fig.~\ref{ts}, le type de GV est $\langle e,t\rangle$. On peut en conclure que $\Trad(\text{GV})$ nécessite un $\lambda$\yh-opérateur, qui doit capter le $e$ pour en faire un $t$. Et c'est ce $\lambda$\yh-opérateur qui va recevoir le $g$ quand on va appliquer $\Trad(\text{GV})$ à $g$. Écrivons donc $\Trad(\text{GV})=\lambda y.\mathrm{aime}(a)(y)$.

Vérifions : $\Trad(\text{GV})(g)=(\lambda y.\mathrm{aime}(a)(y))(g)=\mathrm{aime}(a)(g)=\Trad(\text S)$, donc tout va bien.

De la même manière, on définit $\Trad(\text{V})=\lambda x.\lambda y.\mathrm{aime}(x)(y)$, et on vérifie : $\Trad(\text{V})(a)=(\lambda x.\lambda y.\mathrm{aime}(x)(y))(a)=\lambda y.\mathrm{aime}(a)(y)=\Trad(\text{GV})$. \textsc{cqfd}.

Le lecteur trouvera sur la fig.~\ref{lc} la traduction dans \logique{} de l'arbre syntaxique de la phrase \mysf{Gérard aime Alice} :

\begin{figure}[h]
\Tree [.{$\mathrm{aime}(a)(g)$} [.$g$ [.$g$ $g$ ] ] [.{$\lambda y.\mathrm{aime}(a)(y)$} [.{$\lambda x.\lambda y.\mathrm{aime}(x)(y)$} {$\lambda x.\lambda y.\mathrm{aime}(x)(y)$} ] [.$a$ $a$ ]  ] ]
\caption{Traduction en \logique{} de l'arbre de la fig.~\ref{arbre}.\label{lc}}
\end{figure}

On constate que la taille et la complexité des formules logiques sont assez variées : alors que $\Trad(\text{GN}_1)=g$ et $\Trad(\text{GN}_2)=a$ sont très simples, $\Trad(\text{V})$ est bien plus complexe. On peut dire qu'intuitivement cela montre le potentiel d'action caché dans le verbe : lorsque on monte vers la racine, le verbe agit sur les autres constituants, jusqu'à fournir la phrase complète.

\subsection{Un exemple : la coordination}

Pour montrer la force de cette théorie et exercer un peu nos neurones, posons-nous un petit casse-tête : la coordination.

\begin{figure}[ht]
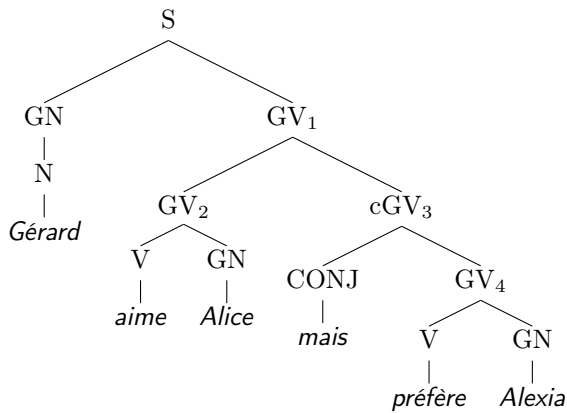

\Tree [.S [.GN [.N \mysf{Gérard} ] ] [.GV$_1$ [.GV$_2$ [.V  \mysf{aime} ] [.GN \mysf{Alice} ] ] [.cGV$_3$ [.CONJ \mysf{mais} ] [.GV$_4$ [.V \mysf{préfère} ] [.GN \mysf{Alexia} ] ] ] ] ]
\caption{Arbre syntaxique de la phrase \mysf{Gérard aime Alice mais préfère Alexia}.\label{arbre2}}
\end{figure}

Le problème commence quand l'insatiable \mysf{Gérard} se met à préférer \mysf{Alexia} alors qu'il aime \mysf{Alice}. Prenons la phrase \mysf{Gérard aime Alice mais préfère Alexia}, et son arbre syntaxique (fig.~\ref{arbre2}). Les indices des GV et cGV servent uniquement à les distinguer dans la suite. On a noté cGV$_3$ la partie \mysf{mais préfère Alexia}, qui n'est pas un simple groupe verbal mais un «groupe verbal muni d'une conjonction» (d'où le «c» de cGV).

Que dire de cette phrase ? Le mot \mysf{mais} correspond logiquement à une conjonction et donc la traduction de S sera $\mathrm{aime}(a)(g)\wedge\mathrm{pr\acute{e}f\grave{e}re}(a')(g)$ où $a= \Trad(\text{\mysf{Alice}})$ et $a'= \Trad(\text{\mysf{Alexia}})$. D'après la section précédente, \Trad(GV$_2$) et \Trad(GV$_4$) seront resp.\ $\lambda y.\mathrm{aime}(a)(y)$ et $\lambda y.\mathrm{pr\acute{e}f\grave{e}re}(a')(y)$.

La grande question est : que seront \Trad(GV$_1$), \Trad(cGV$_3$), et surtout \Trad(CONJ) ?

\begin{figure}[ht]
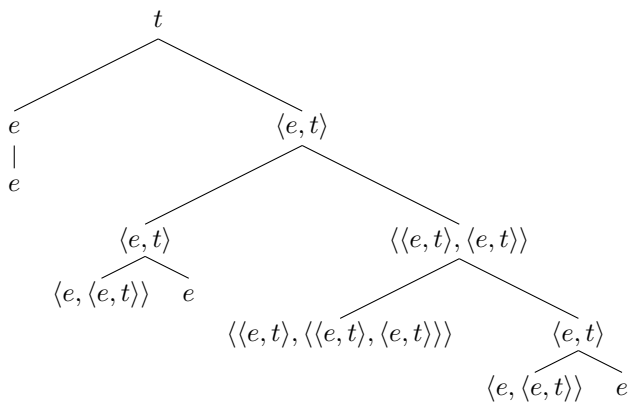

{
\Tree [.$t$ 
[.$e$ $e$ ]
[.$\langle e,t\rangle$
   [.$\langle e,t\rangle$
      [.$\langle e,\langle e,t\rangle\rangle$ ]
      [.$e$ ]
   ]
   [.$\langle\langle e,t\rangle,\langle e,t\rangle\rangle$  
      [.$\langle\langle e,t\rangle,\langle\langle e,t\rangle,\langle e,t\rangle\rangle\rangle$ ] 
      [.$\langle e,t\rangle$ 
      [.$\langle e,\langle e,t\rangle\rangle$ ] [.$e$ ] ]
   ]
]
]}
\caption{Arbre des types de la phrase \mysf{Gérard aime Alice mais préfère Alexia}.\label{ts2}}
\end{figure}

Traçons d'abord l'arbre des types (fig.~\ref{ts2}). Pour GV$_1$, GV$_2$, GV$_4$, rien de nouveau, ce sont des $\langle e,t\rangle$, comme tout GV qui se respecte. Quid de cGV$_3$ ? Coincé entre GV$_1$ et GV$_2$, il ne peut être que $\langle\langle e,t\rangle,\langle e,t\rangle\rangle$. Et donc, CONJ ne peut être que $\langle\langle e,t\rangle,\langle\langle e,t\rangle,\langle e,t\rangle\rangle\rangle$ (intuitivement : à partir d'un prédicat, et ensuite d'un deuxième prédicat, il fournit un nouveau prédicat). 

Et maintenant, pour trouver \Trad(GV$_1$), \Trad(cGV$_3$) et  \Trad(CONJ), allons de nouveau à reculons :

1. Pour trouver \Trad(GV$_1$) il faut éliminer (le terme correct est «\emph{$\lambda$\yh-abstraire}») $g$ de $\Trad(\text{S})=\mathrm{aime}(a)(g)\wedge\mathrm{pr\acute{e}f\grave{e}re}(a')(g)$. On le fait en écrivant $\Trad(\text{S})=\lambda x.(\mathrm{aime}(a)(x)\wedge\mathrm{pr\acute{e}f\grave{e}re}(a')(x))(g)$ et donc $\Trad(\text{GV}_1)=\lambda x.(\mathrm{aime}(a)(x)\wedge\mathrm{pr\acute{e}f\grave{e}re}(a')(x))$.

2. Essayons de $\lambda$\yh-abstraire $\Trad(\text{GV}_2)$ de $\Trad(\text{GV}_1)$ pour obtenir $\Trad(\text{cGV}_3)$ :%
\break
on a $\Trad(\text{GV}_1)=\lambda x.(\mathrm{aime}(a)(x)\wedge\mathrm{pr\acute{e}f\grave{e}re}(a')(x))=\lambda x.\mathrm{aime}(a)(x)\wedge\lambda x.\mathrm{pr\acute{e}f\grave{e}re}(a')(x)$. Le terme $\lambda x.\mathrm{aime}(a)(x)$ est un prédicat.

On peut le $\lambda$\yh-abstraire en l'écrivant sous la forme 
$\lambda P.(P)(\lambda x.\mathrm{aime}(a)(x))$,
ce qui donne%
\break
$\Trad(\text{GV}_1)=\lambda P.(P\wedge\lambda x.\mathrm{pr\acute{e}f\grave{e}re}(a')(x))(\lambda x.\mathrm{aime}(a)(x))$ et donc, par le principe de composition-application, $\Trad(\text{cGV}_3)=\lambda P.(P\wedge\lambda x.\mathrm{pr\acute{e}f\grave{e}re}(a')(x))$.

3. De la même manière, nous $\lambda$\yh-abstrayons $\Trad(\text{GV}_4)$ de $\Trad(\text{cGV}_3)$ et ce qui reste sera \Trad(CONJ). On trouve $\Trad(\text{cGV}_3)=\lambda Q.\lambda P.(P\wedge Q)(\lambda x.\mathrm{pr\acute{e}f\grave{e}re}(a')(x))$ et donc, enfin,%
\break
$\Trad(\text{CONJ})=\lambda Q.\lambda P.(P\wedge Q)$.

À y réfléchir, ce résultat n'a rien d'étonnant : après tout, un opérateur binaire comme $\wedge$ n'associe-t-il pas deux prédicats $P$ et $Q$ au prédicat $P\wedge Q$ ?

\subsection{La quantification}

Prenons maintenant la phrase \mysf{tout le monde aime Alice}. La différence avec \mysf{Gérard aime Alice} est que si \mysf{Gérard} (grammaticalement, un nom) peut être traduit par une constante $g$ et puis interprété par une entité \textbf{gérard}, on ne peut faire de même pour \mysf{tout le monde} (un pronom indéfini), que l'on sera logiquement obligé d'interpréter par la totalité des objets de \monde{} (puisque c'est «tout le monde» qui aime \mysf{Alice}).

\begin{figure}[ht]
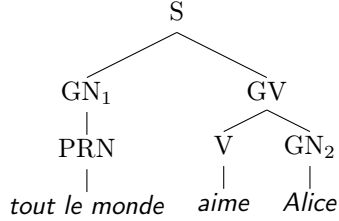

\Tree [.S [.GN$_1$ [.PRN \mysf{tout le monde} ] ] [.GV [.V \mysf{aime} ] [.GN$_2$ \mysf{Alice} ] ] ]
\caption{Arbre syntaxique de la phrase \mysf{tout le monde aime Alice}.\label{quant}}
\end{figure}

\begin{figure}[ht]
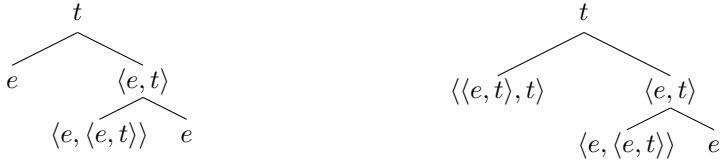

\Tree [.{$t$} [.$e$ ] [.$\langle e,t\rangle$ [.{$\langle e,\langle e,t\rangle\rangle$} ] [.$e$ ]  ] ]\qquad
\Tree [.$t$ [.$\langle \langle e,t\rangle ,t\rangle $ ] [.$\langle e,t\rangle $ [.$\langle e,\langle e,t\rangle \rangle $ ] [.$e$ ] ] ]
\caption{Arbres des types des phrases (a) \mysf{Gérard aime Alice} et (b) \mysf{tout le monde aime Alice}.\label{tquant}}
\end{figure}

Cela ne va pas changer outre mesure l'arbre syntaxique de la phrase (fig.~\ref{quant}), mais on voit la différence au niveau de l'arbre des types (fig.~\ref{tquant}b). Là où \mysf{Gérard} était traduit par un type $e$, \mysf{tout le monde} est traduit par un $\langle \langle e,t\rangle ,t\rangle $, c'est-à-dire qu'il prend le groupe verbal $\lambda x.\mathrm{aime}(a)(x)$ en entrée et retourne une valeur de vérité (la réponse à la question : «tout le monde aime-t-il \mysf{Alice} ?»).

Quelle va être la traduction de S ? Il est naturel de se servir du quantificateur universel pour écrire $\Trad(S)=\forall x\,(\mathrm{aime}(a)(x))$. 

Sachant que GV se traduit par $\lambda x.\mathrm{aime}(a)(x)$, quelle va être la traduction de PRN ? Voici comment s'y prendre : posons $P=\lambda x.\mathrm{aime}(a)(x)$, $P$ est donc un prédicat unaire. S devient alors $\forall x\,(P(x))$, que l'on peut $\lambda$\yh-abstraire en $(\lambda Q.(\forall x\,(Q(x)))(P)$, et on reconnaît ici une fonction de prédicat appliquée au prédicat $P$. Mais ce prédicat n'est autre que la traduction de GV, et on a donc trouvé une fonction qui, appliquée à GV, nous donne S : d'après le principe de compositionnalité, cela n'est rien d'autre que la traduction de PRN.

On remplace donc P par sa valeur et on a $\Trad(\text{\mysf{tout le monde}})=\lambda P.(\forall x\,(P(x)))$, qui est bien de type $\langle \langle e,t\rangle ,t\rangle $.

\subsection{L'article défini}

Que de plus simple dans la langue française que l'article défini «le, la, les» ? Et pourtant, sa traduction en \logique{} sera pour nous un petit challenge !

\begin{figure}[ht]
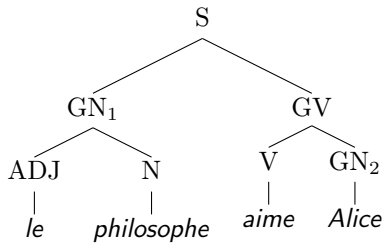

\Tree [.S [.GN$_1$ [.ADJ \mysf{le} ] [.N \mysf{philosophe} ] ] [.GV [.V \mysf{aime} ] [.GN$_2$ \mysf{Alice} ] ] ]
\caption{Arbre syntaxique de la phrase \mysf{le philosophe aime Alice}.\label{artdef}}
\end{figure}

Prenons la phrase \mysf{le philosophe aime Alice} (fig.~\ref{artdef}). Notons tout de suite que \mysf{philosophe} ne peut être traduit par une constante, comme, par exemple, \mysf{Gérard}, puisque «être philosophe» est une propriété, et les propriétés sont traduites par des prédicats unaires. Ainsi, on écrira $\mathrm{philosophe}(g)$ pour dire que $g$ est philosophe. De même, $\exists x\,\mathrm{philosophe}(x)$ signifie qu'il existe un philosophe. Et donc, si la phrase de départ était \mysf{\textbf{un} philosophe aime Alice} (avec un article indéfini), sa traduction en \logique{} serait $\exists x\,(\mathrm{philosophe}(x)\wedge\mathrm{aime}(a)(x))$.

\begin{figure}[ht]
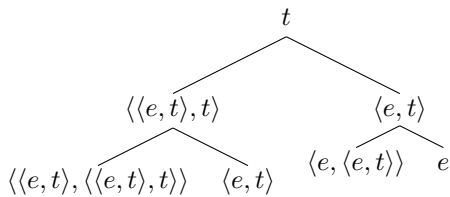

\Tree [.$t$ [.$\langle \langle e,t\rangle ,t\rangle $ [.$\langle \langle e,t\rangle ,\langle \langle e,t\rangle ,t\rangle \rangle $ ] [.$\langle e,t\rangle $ ] ] [.$\langle e,t\rangle $ [.$\langle e,\langle e,t\rangle \rangle $ ] [.$e$ ] ] ]
\caption{Arbre des types de la phrase \mysf{le philosophe aime Alice}.\label{tartdef}}
\end{figure}

D'où l'arbre de types de la phrase \mysf{le philosophe aime Alice} (fig.~\ref{tartdef}), où l'on affecte à \mysf{philosophe} le type $\langle e,t\rangle $ (le même que celui du verbe \mysf{aime}) et donc GN$_1$ devient $\langle \langle e,t\rangle ,t\rangle $. Il ne nous reste d'autre choix pour \mysf{le} que de lui affecter le type $\langle \langle e,t\rangle ,\langle \langle e,t\rangle ,t\rangle \rangle $, autrement dit : il prend un $\langle e,t\rangle $ (\mysf{philosophe}) et ensuite un autre $\langle e,t\rangle $ (\mysf{aime}) et retourne une valeur de vérité.

Mais comment traduire alors l'article défini \mysf{le} ?

Formulons la question autrement : comment indiquer qu'il n'y a qu'\emph{un seul} philosophe, et que quand on dit \mysf{le philosophe} on parle justement de lui ?

Pour répondre à cette question, rappelons-nous que dans la définition de la formule logique de 1\up{er} ordre (déf.~\ref{fl}, p.~\pageref{fl}) nous avons mentionné la relation binaire~$=$ («égalité»). Mais qu'est-ce donc l'«égalité» ?

C'est une vaste question philosophique dont on ne parle guère en cours de mathématiques... et pourtant, en logique du 1\up{er} ordre, l'«égalité» a un sens bien particulier, que voici : \emph{écrire $a=b$ signifie que l'on demande que dans toute interprétation, les constantes (ou variables) $a$ et $b$ soient interprétées par le même élément de \monde}. 

Et c'est ainsi qu'on traduit l'\emph{unicité} : il n'y a qu'un seul philosophe $x$ si et seulement si pour tout individu $y$ tel que $y$ soit philosophe, on ait $x=y$.

La phrase \mysf{le philosophe aime Alice} se traduira donc par $\exists x\,(\forall y\,(\mathrm{philosophe}(y) \leftrightarrow x=y)\wedge\mathrm{aime}(a)(x))$.

En appliquant les mêmes méthodes de $\lambda$\yh-abstraction que dans la section précédente, on trouve que la traduction de l'article défini \mysf{le} ne peut être que $$\Trad(\text{\mysf{le}})=\lambda Q.(\lambda P.(\exists x\,(\forall y\,(Q(y)\leftrightarrow (x=y))\wedge P(x)))).$$
Le lecteur peut imaginer le désarroi de l'étudiant en linguistique qui, ayant raté le premier cours, se retrouve devant une formalisation de la langue naturelle qui traduit un des mots les plus simples de sa langue par cette horrible formule... Est-ce bien raisonnable ? En fait, nous n'avons fait que rendre visible l'important potentiel sémantique de ce petit mot grammatical --- à première vue, insignifiant --- qu'est l'article défini.

\section{L'ensemble \monde{} et la fonction d'interprétation \Inter}\label{monde}

Jusqu'ici nous avons surtout parlé de $\Trad:\langue{}\to\logique{}$. Il ne reste plus qu'à décrire plus précisément \monde{}, ainsi que la fonction $\Inter:\logique\to\monde$. Pour cela, nous allons nous servir d'un autre outil mathématique, bien plus connu cette fois-ci : la \emph{théorie des ensembles}.

Ainsi, si $a=\Trad(\text{\mysf{Alice}})$ et $g=\Trad(\text{\mysf{Gérard}})$ deviennent dans \monde{} la belle \textbf{alice} et le vilain \textbf{gérard}, que dire alors des autres sommets des arbres syntaxiques que nous avons étudiés ?

Là aussi Montague a eu les bonnes idées ! 

D'après la définition de l'interprétation d'une formule (déf.~\ref{inter}), un prédicat unaire est interprété par une fonction $\monde\to Z_2$, où $Z_2$ est l'ensemble $\{\mathbf{vrai},\mathbf{faux}\}$. Notons par $Z^\monde_2$ les fonctions de $\monde$ dans~$Z_2$. Alors $\Inter(\lambda x.P(x))\in Z^\monde_2$, c'est-à-dire que l'interprétation d'un prédicat unaire de \logique{} est un élément de $Z^\monde_2$.

Cette propriété se généralise aux prédicats $n$\yh-aires quelconques. En effet, il suffit de constater qu'un prédicat binaire devient une fonction qui à chaque élément de~$\monde$ associe un prédicat unaire. Donc $\Inter(\lambda x.\lambda y.P(x)(y))\in Z^{\monde^\monde}_2$, et ainsi de suite...

Cela semble abstrait, mais en réalité on ne fait que manipuler de simples relations binaires. Prenons un exemple : supposons que $\monde=\{\text{\textbf{alice}},\text{\textbf{gérard}},\text{\textbf{billy}}\}$ (notés dans \logique{} par $a,g,b$) et que l'on ait $\Inter(\mathrm{aime}(a)(g))=\text{\textbf{vrai}}$ et $\Inter(\mathrm{aime}(a)(b))=\text{\textbf{vrai}}$. Que va être $\Inter(\lambda y.\mathrm{aime}(a)(y))$ ? Ce sera un élément de $Z_2^\monde$ : l'ensemble de relations binaires 
$\phi_a=\{
(\text{\textbf{alice}},\text{\textbf{faux}}),(\text{\textbf{gérard}},\text{\textbf{vrai}}),(\text{\textbf{billy}},\text{\textbf{vrai}})\}$ (puisque \textbf{gérard} et \textbf{billy} aiment \textbf{alice}, mais \textbf{alice} ne s'aime pas elle-même). Nous avons appelé $\phi_a$ cet ensemble de relations binaires puisqu'il s'agit de savoir qui aime \textbf{alice}. De la même manière, il existe $\phi_g$ et $\phi_b$ qui concernent \textbf{gérard} et \textbf{billy}.

Que sera alors $\Inter(\lambda x.\lambda y.\mathrm{aime}(x)(y))$ ? Ce sera un élément de $Z_2^{\monde^\monde}$ : un ensemble de relations binaires entre des éléments de $\monde$ et des fonctions $\monde\to Z_2$. Dans notre cas, ce sera tout simplement $\Phi=\{(\text{\textbf{alice}},\phi_a),(\text{\textbf{gérard}},\phi_g),(\text{\textbf{billy}},\phi_b)\}$.

Supposons que nous souhaitions savoir si $\Inter(\mathrm{aime}(g)(a))$ est $\text{\textbf{vrai}}$. Écrivons \begin{align*}\Inter&(\mathrm{aime}(g)(a))\\&=
\Inter((\lambda x.\lambda y.\mathrm{aime}(x)(y))(g)(a))\\&=
\Inter(\lambda x.\lambda y.\mathrm{aime}(x)(y))(\Inter(g))(\Inter(a))\\&=
\Phi(\Inter(g))(\Inter(a))\\&=
\phi_g(\Inter(a))=\text{\textbf{faux}},\end{align*}
donc, hélas, \textbf{alice} n'aime (toujours) pas \textbf{gérard}.

On voit donc de quelle manière il est possible d'interpréter n'importe quelle formule de \logique.

\section{Et le reste...}

Cet article est déjà assez long. Jusqu'ici, nous avons parcouru et partiellement illustré une partie de la théorie de Montague. Le restant de sa théorie est tout aussi utile et intéressant, mais n'implique pas de nouvel outil mathématique ; nous allons donc nous contenter d'en décrire rapidement les grandes lignes.

\subsection{Inférence}

Ce qui fait la force de la logique mathématique est le fait qu'à partir d'un ensemble de formules (que l'on considère vraies pour une interprétation donnée), on a des mécanismes (appelés \emph{règles d'inférence}) pour obtenir de nouvelles formules (également vraies dans la même interprétation). On admet donc certaines formules en tant qu'\emph{axiomes} et on en déduit d'autres, appelées \emph{théorèmes}. C'est ainsi que fonctionnent les mathématiques : que ce soit en géométrie, en algèbre ou en analyse, on admet certains axiomes et on construit des \emph{théories} en démontrant des \emph{théorèmes}.

Un exemple de règle d'inférence très utile est le \emph{modus ponens} : si on a $\phi\to\psi$ et $\phi$ alors on peut en déduire $\psi$ (exemple : si on admet que \mysf{tous les hommes sont mortels} et que \mysf{Socrate est un homme}, alors on peut en déduire que \mysf{Socrate est mortel}).

Montague introduit le mécanisme des règles d'inférence dans \logique{}.

\subsection{La temporalité}

Pour analyser des phrases comme \mysf{Gérard n'aime plus Alice}, Montague introduit la temporalité dans \logique. Le temps y est représenté de deux manières : par des \emph{instants} et par des \emph{intervalles temporels}. Il propose des opérateurs entre les intervalles : deux intervalles $[t_1,t_2]$ et $[t_3,t_4]$ peuvent se chevaucher (quand, par exemple, $t_1<t_3<t_2<t_4$), être totalement disjoints ($t_2<t_3$) ou s'imbriquer l'un dans l'autre ($t_1<t_2<t_3<t_4$).

Chaque formule de \logique{} est indexée temporellement : sa valeur de vérité dépend (outre les variables libres qu'elle contient) de ses propriétés temporelles.

\subsection{Les mondes possibles/accessibles}\label{mp}

Pour analyser des phrases comme \mysf{Gérard aime peut-être Alice, mais certainement pas Alexia}, Montague introduit les notions de \emph{modalité} et de \emph{mondes accessibles}. Il propose deux opérateurs modaux : $\square \phi$ qui signifie «$\phi$ est nécessairement vraie», et $\diamond \phi$ qui signifie «$\phi$ est peut-être vraie».

Pour formaliser ces notions, il parle de «mondes accessibles» : en effet, on a toujours dit que \monde{} pouvait être un monde hypothétique quelconque, alors pourquoi ne pas en imaginer plusieurs, voire \emph{tous} les mondes possibles et imaginables ? Mais comme cela est un peu éloigné de la réalité des problèmes que l'on peut se poser, il est plus raisonnable de parler de mondes «accessibles» : ce sont ceux qui constituent des alternatives plausibles à un monde donné. Si $\monde$ et $\monde'$ sont des mondes, on peut imaginer une relation binaire $\mathrm{Accessible}(\monde,\monde')$ qui signifie que $\monde'$ est accessible à partir de $\monde$.

Alors la formule $\phi$ est «nécessairement vraie» ($\square\phi$) si elle est vraie dans \emph{tous} les mondes accessibles à partir du monde courant, et elle est «peut-être» vraie ($\diamond\phi$), si elle est vraie dans \emph{certains} de ces mondes.

\subsection{Intensionnalité}

Définir un ensemble \emph{extensionnellement} consiste à énumérer ses éléments, le définir \emph{intensionnellement} consiste à en donner les propriétés. Ainsi, ${\{p\mid p\leq 10,p\text{ premier}\}}$ est une définition intensionnelle et $\{2,3,5,7\}$ la définition sensationnelle du même ensemble.

Montague se sert de l'intensionnalité pour englober en un seul objet mathématique les valeurs d'une expression dans tous les mondes accessibles. Ainsi, si $\mathrm{aime}(a)(g)$ est vrai dans les mondes $\monde_1$, $\monde_2$ et $\monde_4$, et $\mathrm{aime}(a')(g)$ est vrai dans les mondes $\monde_1$, $\monde_2$ et $\monde_3$, alors ces deux formules ont la même valeur de vérité dans \emph{certains} mondes mais pas dans \emph{tous}. On dira que leurs \emph{intensions} sont différentes.

L'\emph{intension d'une formule} est donc une fonction qui envoie différents mondes vers les valeurs de vérité correspondantes des formules. En comparant les intensions de deux formules, on compare leur «comportement» dans tous les mondes accessibles, c'est bien plus puissant que de les comparer dans un seul monde. Cette notion est tellement importante pour Montague que sa théorie est souvent appelée \emph{sémantique intensionnelle}.

\section{Conclusion}

Nous avons exploré les outils mathématiques qui ont servi à Montague et à ceux qui l'ont suivi pour modéliser la langue naturelle en tant que langage formel. Ce parcours nous a permis d'évoquer et de décrire brièvement les langages et grammaires formels, une version légèrement restreinte de la logique de 1\up{er} ordre, la théorie de types, le $\lambda$\yh-calcul. Dans chaque cas nous avons tenté de justifier l'utilité de l'outil pour l'analyse de la langue naturelle, et de l'illustrer par des exemples.

Notre but a été de faire découvrir au lecteur comment les mathématiques permettent d'étudier les liens entre la langue naturelle et le monde --- liens qui nous affectent tous profondément puisque, comme disait Wittgestein :
\begin{quote}«\emph{die Grenzen meiner Sprache bedeuten die Grenzen meiner Welt}»\\
(= les limites de ma langue sont les limites de mon propre monde) \cite[\S\,5.6]{witt}\end{quote} 

\subsection*{Conseils de lecture}

Les langages et grammaires formels sont décrits avec beaucoup de rigueur dans~\cite{dehor}. La logique du 1\up{er} ordre est admirablement bien présentée par J.-P.~Delahaye dans~\cite{delahaye}.

En ce qui concerne la sémantique de Montague, une présentation très accessible est donnée dans~\cite{gal}. L'ouvrage \cite{cham} est un peu plus technique mais tout aussi abordable. Enfin, \cite{gochet}, tiré du Séminaire de philosophie et mathématiques de l'ENS (dirigé, excusez du peu, par J.~Dieudonné et R.~Thom), en donne une synthèse très efficace. Dans son exposé filmé «La modélisation mathématique des langues naturelles» de l'Université de tous les savoirs, S.~Kahane parle de la sémantique de Montague \cite[$30'24''\text{--}34'45''$]{kaha1} avant d'enchaîner sur une de ses propres contributions : les grammaires à bulles~\cite{kaha2}.

Enfin, en ce qui concerne la vie (et la mort) de Richard~M. Montague, voir l'essai \emph{That's just Semantics!} \cite{web}, ainsi que les romans \cite{ber}, \cite{meurtre} et \cite{delany}.

\bibliographystyle{plain}
\bibliography{main}

\vfill

\end{document}